\pdfoutput=1
\documentclass[11pt]{article}
\usepackage[]{acl}
\usepackage{times}
\usepackage{hyperref}
\usepackage{latexsym}
\usepackage{multirow}
\usepackage{graphicx}
\usepackage{amsmath}
\usepackage{amssymb}
\usepackage{booktabs}
\usepackage{subcaption}
\usepackage{caption}
\usepackage{bbding}
\usepackage{colortbl}
\usepackage{tcolorbox}
\usepackage{pifont}
\usepackage{algpseudocode}
\usepackage{algorithm}
\newcommand{\ttt}{\texttt}
\newcommand{\ie}{\textit{i.e.}}
\newcommand{\eg}{\textit{e.g.}}
\newtcolorbox[list inside=prompt,auto counter]{prompt}[1][]{
    colbacktitle=black!60,
    coltitle=white,
    fontupper=\footnotesize,
    boxsep=5pt,
    left=0pt,
    right=0pt,
    top=0pt,
    bottom=0pt,
    boxrule=1pt,
    #1,
}
\usepackage[T1]{fontenc}
\usepackage[utf8]{inputenc}
\usepackage{microtype}
\usepackage{inconsolata}
\title{UFO: a Unified and Flexible Framework for Evaluating Factuality of Large Language Models}
\author{Zhaoheng Huang$^1$, Zhicheng Dou$^{1}$\thanks{$^*$Corresponding author.}, Yutao Zhu$^1$, \and Ji-rong Wen$^1$\\
$^1$Gaoling School of Artificial Intelligence, Renmin University of China, Beijing, China \\
\texttt{\{huangzh,dou,ytzhu,jrwen\}@ruc.edu.cn}  
}
\newcommand{\ours}{\ttt{UFO} }
\newcommand{\chatgpt}{\ttt{gpt-3.5-turbo-1106}}
\def\debug{}  
\ifdefined\debug

\else

\fi
\begin{document}
\maketitle

\begin{abstract}
% 1. 研究背景
% 2. 现有工作研究
% 3. 现有工作缺点
% 4. 我们为了什么而提出了什么
% 5. 我们的实验和结论
% However, obtaining annotated data for language models is difficult and costly.
% Moreover, these metrics overlook reference documents provided by some LLMs and do not effectively verify open-domain facts in the generated text.
% However, the reliability of human-annotated evidence and reference documents, which are used as different fact sources for the given question, remains under-explored. The availability of fact sources varies according to the nature of the generative task, forming different evaluation scenarios. Moreover, the effectiveness of verification in scenarios where certain LLM-based fact extractors are unavailable has not been adequately examined. 
% To address these challenges, we categorized two types of available fact sources 
% and four evaluation scenarios containing six representative datasets. Further, we propose \ours, an LLM-based unified and flexible evaluation pipeline to verify facts against plug-and-play fact sources.
% Further, we propose \ours, a unified and flexible evaluation pipeline with multi-level available fact sources to align the human evaluation process. Our proposed pipeline verifies each fact unit in the model-generated text at three different levels.
% Experimental results demonstrate that the two types of fact sources are crucial in the verification of most QA tasks, while in News Fact Generation and retrieval-augmented QA tasks, they either negatively affect the outcomes or can be substituted for one another.
% Our dataset and code are available at \url{https://github.com/WaldenRUC/UFO}.
Large language models (LLMs) may generate text that lacks consistency with human knowledge, leading to factual inaccuracies or \textit{hallucination}.
Existing research for evaluating the factuality of LLMs involves extracting fact claims using an LLM and verifying them against a predefined fact source.
However, these evaluation metrics are task-specific, and not scalable, and the substitutability of fact sources in different tasks is under-explored.
To address these challenges, 
we categorize four available fact sources: human-written evidence, reference documents, search engine results, and LLM knowledge, along with five text generation tasks containing six representative datasets. Then, we propose \texttt{UFO}, an LLM-based unified and flexible evaluation framework to verify facts against plug-and-play fact sources. We implement five evaluation scenarios based on this framework.
Experimental results show that for most QA tasks, human-written evidence and reference documents are crucial, and they can substitute for each other in retrieval-augmented QA tasks. In news fact generation tasks, search engine results and LLM knowledge are essential. 
Our dataset and code are available at \url{https://github.com/WaldenRUC/UFO}.
\end{abstract}

\section{Introduction}
% 1. 语言模型的发展与幻觉
The advancement of large language models (LLMs) has facilitated the development of generative artificial intelligence~\cite{LLMsurvey}. Many LLM-based applications have been released, such as ChatGPT and Bing Chat (also known as Bing Copilot), which gradually change people's working habits.\footnote{ChatGPT: \url{https://chat.openai.com/chat}, Bing Chat: \url{https://www.bing.com/new}} 
However, LLMs tend to generate factually inaccurate texts, which lack consistency with human knowledge, and degrade the usability of the model-generated text. Such a shortcoming of LLMs is well-known as \textit{hallucination}~\cite{Hallucination, Hallucination2}.
% 2. 产生幻觉的原因，介绍现有的评测方式(ref-based, ref-free)，引入事实源的定义
The quality of datasets and training paradigms are concerned as the potential factors causing hallucinations in LLMs~\cite{faithfulnessSurvey}. How to detect and measure the hallucinations in model-generated texts has received increasing attention. 
% A common practice in existing automatic factuality evaluation metrics is to compare the fact consistency between the model-generated text and the \textit{human-written evidence}, where the latter is used as the \textbf{fact source}~\cite{QAGS, Q2, TripleScore}. Meanwhile, some reference-free evaluation metrics rely solely on the web corpus as the fact source to verify each atomic fact unit in the model-generated text~\cite{factscore, factool}.

\begin{figure}[tb]
    \centering
    \includegraphics[width=1\linewidth]{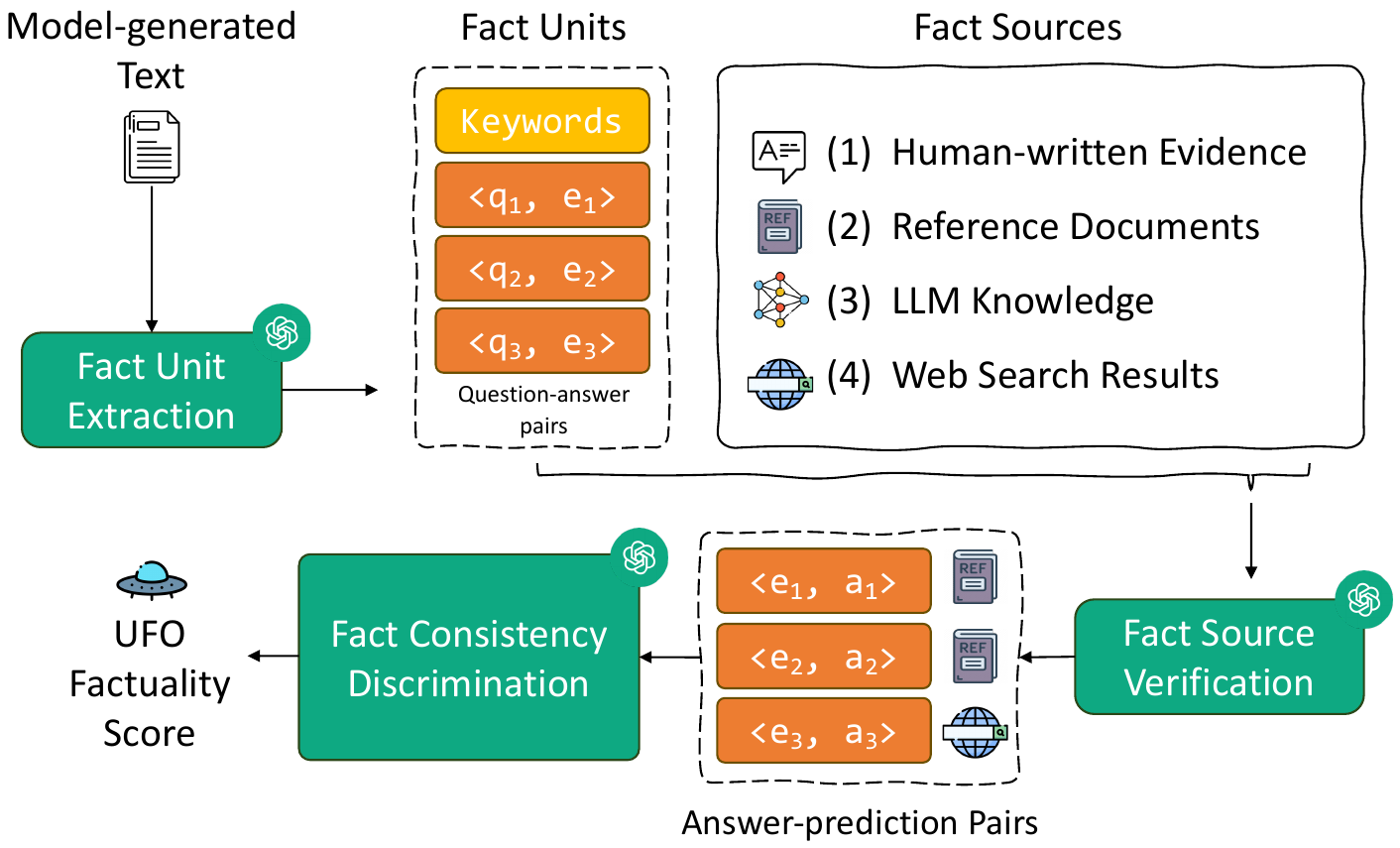}
    \caption{Our proposed factuality evaluation pipeline \ours. We integrate four fact sources within various evaluation scenarios to assess the factuality score.}
    \label{fig:overallArc}
\end{figure}

% 3. 现有工作的问题: 
% Such automatic evaluation metrics have two limitations:  
% 3-1. 这些评测只限于特定的任务，扩展性不足，当我们设立一个新的数据集评测大模型的事实准确性，我们需要探索是否有必要为这个数据集搜集标注数据。
% (1) These evaluations are limited to specific tasks with insufficient scalability. When establishing a new dataset to evaluate hallucinations in LLMs, it is necessary to determine the necessity of collecting annotations as the fact source for this dataset. 
% It is also essential to explore scenarios where the facts in the model-generated text may not be verified by predefined fact sources, necessitating the need for annotations.
% % 3-2. 这些模型在区分不同模型的事实准确度上(Discriminative Power)存在不足
% (2) These evaluation metrics have limited capacity to distinguish the factual accuracy of different model-generated text, and thus are not effective in accurately assessing the differences in factual accuracy between two LLMs.
% 修改后的表述: 强调了评估工作的unified和flexible的必要性
% 介绍已有的方法, 按照事实源的脉络介绍，但不提事实源的事情，只说方法使用什么检测的, 然后说我们的方法里考虑了哪四种事实源，也不需要太细，描述一下就可以了，然后说这些事实源的组合构成了哪些典型的应用场景，并cite table1
Current automatic evaluation metrics employ a specific fact source to evaluate the factuality of LLMs for certain tasks. 
However, there is still a lack of analysis on the applicability of different fact sources in various tasks.
Considering the establishment of a new task, the fact sources relied upon by previous evaluation methods may not be applicable. It's important to consider whether alternative fact sources can be utilized.
For example, when a new QA task arises, collecting human-written evidence can be extremely costly. In such cases, whether search results from a search engine can be used as a substitute for human-written evidence as a fact source remains unexplored.

To address the issue, we propose \ours, a \textbf{U}nified and \textbf{F}lexible framework for factuality evaluati\textbf{O}n, which allows for:
(a) Flexibly integration of various fact sources.
(b) A unified verification method that enables switching fact sources in specific tasks.
(c) The combination of different fact sources to enhance the factuality evaluation.
In our framework shown in Figure~\ref{fig:overallArc}, we first extract fact units from the text, including verifiable question-answer pairs and keywords of model-generated text. 
Then, for each fact, we verify it against the set of fact sources until a matching answer is found.
% Then we sequentially verify each fact unit by extracting answers from text passages in the fact sources of the certain evaluation scenario. 
Finally, we assign a binary matching score to each fact.

With the support of this evaluation framework, we can systematically analyze the evaluation capabilities of different fact sources across various scenarios in existing evaluation tasks.
Specifically, we consider four different fact sources: 
\textbf{(1) Human-written evidence}. This corresponds to some text generation tasks with labeled data. For example, expert-validated QA tasks often provide human-written answers for evaluation. 
\textbf{(2) Reference documents}. Many recent studies, \eg, WebBrain~\cite{webbrain}, WebGPT~\cite{webgpt}, GopherCite~\cite{gophercite}, WebCPM~\cite{webcpm}, WebGLM~\cite{webglm}, ALCE~\cite{LLMCitation} and Bing Chat, have reported that leveraging reference documents can facilitate LLMs generation of more factual text. Therefore, such reference documents can also be a fact source for factuality evaluation. 
\textbf{(3) Search engine results}. When humans are asked to check the factuality of a text, they usually make judgments by turning to search engines. 
\textbf{(4) LLM knowledge}. Existing studies~\cite{llmasjudges} suggest that advanced LLMs (such as GPT-4) can serve as a fact source for verification.
We design five evaluation scenarios where different fact sources and their combinations are used, summarized in Table~\ref{tab:scenarios}, to demonstrate the flexibility of \ours.
%(NQ~\cite{NQopen}, HotpotQA~\cite{hotpotQA}, TruthfulQA~\cite{truthfulQA}, CNN/Daily Mails~\cite{cnndm}, Multi-News~\cite{multinews}, and MS MARCO~\cite{msmarco}).
%We design five evaluation scenarios based on fact source availability and verification order in the dataset to analyze their importance in different tasks.
%
% 在每个scenario下，我们均计算DP，并与baseline对比，实验表明在大部分QA任务上，he和rd事实源的标注很重要；在新闻事实生成任务上，se和lk事实源即可表现很好的判别力；在RAQA任务上，获取he和rd其中之一即可提升判别力。
% In each evaluation scenario, we compute the discriminative power~\cite{dp} of our proposed framework and compare it with a series of baseline metrics (BLEU~\cite{bleu}, ROUGE~\cite{ROUGE}, BERTScore~\cite{Bertscore}, BARTScore~\cite{bartscore}, QAGS~\cite{QAGS}, $Q^2$~\cite{Q2}, FactScore~\cite{factscore}, and FacTool~\cite{factool}).
In each evaluation scenario, we compute the discriminative power~\cite{dp} of our proposed framework and compare it with eight baseline metrics.
We experiment with these evaluation scenarios over five text-generation tasks, including Open-domain QA, Web Retrieval-based QA, Expert-Validated QA, News Fact Generation, and Retrieval-Augmented QA, to investigate the importance of data sources in different task scenarios. The experimental results demonstrated that in most QA tasks, obtaining human-written evidence and reference documents enhances the discriminative power of the evaluation pipeline. 
In the news fact generation task, we only require the search engine results and LLM knowledge to verify facts.
In the retrieval-augmented QA task, the positive effects derived from two fact sources are comparable, thus allowing them to be substituted for each other.
% 补一下第四点
Although not the main focus of this paper, we evaluate six existing LLMs: Bing Chat in ``precise'' generation mode, ChatGPT, LLaMA-7b, LLaMA-13b, Vicuna-7b, and Vicuna-13b.
We discovered that the factuality score of Bing Chat in precise mode is lower than that of ChatGPT, yet comparable to Vicuna-13b. In open-source LLMs, increasing the scale of model parameters can enhance factual accuracy.

% While LLMs are capable of a variety of natural language processing tasks, their performance is not usually comparable to fine-tuned PLMs~\cite{ChatGPTNLP}. Additionally, calling the API of LLMs is time-consuming and costly. 
% Existing work demonstrates that LLMs can serve as a knowledge base and assist humans in the fact verification process~\cite{LLMJudge}.
% Thus, we use the PLMs to extract fact units from the given model-generated text and compare the consistency of the facts with available fact sources, and use the LLMs and the search engine to serve as the fact sources for verification.
% We first verify each fact unit with the fact source GT, since the human-annotated text is the most topic-related to the generated text. 
% % 评测流程的简要介绍
% The retrieved reference documents cover certain parts of the aspects of the facts in the model-generated text.
% The fact source GT and RD provide the most relevant and up-to-date factual basis. 
% We then utilize LLMV and SEV to generate and retrieve passages to verify the remaining fact units.
% The final automatic evaluation score is the average of all scores of fact units in the model-generated text.
% The multi-level fact sources are plug-and-play for different generation scenarios.

% 3-6. 本文的主要贡献
% $\bullet$ We categorize existing text generation tasks into four evaluation scenarios and verify model-generated text against provided fact sources.
% $\bullet$ Our experiments have revealed that in certain scenarios and most QA tasks, it is necessary to collect high-quality annotations for verification.
Our contributions can be summarized as follows: 

$\bullet$ We propose \ours, a pipeline integrating flexible plug-and-play fact sources with unified verification methods for evaluating the factuality of LLMs.

$\bullet$ We conduct a systematic analysis of the evaluation capabilities of four fact sources across five factuality evaluation scenarios and five tasks.

$\bullet$ We reveal that human-written evidence and reference documents are essential in QA tasks, while search engine results and LLM knowledge are crucial in news fact generation tasks.

\begin{table}[tb]
    \centering
    \small
    % \caption{In our unified framework \ours, we categorize five text generation tasks into four distinct factuality evaluation scenarios, based on human-written evidence and reference documents.}
    %two types of fact sources in the dataset: human-written evidence and reference documents.}
    \begin{tabular}{p{0.2\linewidth}|p{0.7\linewidth}}
    \toprule
        % \multirow{2}{*}{Tasks}  & \textit{Human-written} & \textit{Reference} \\
        %                             & \textit{Evidence} & \textit{Documents} \\
        % \midrule
        % Open-domain QA                  & \ding{55} & \ding{55} \\
        % Web Retrieval-based QA          & \ding{55} & \ding{51} \\
        % Expert-Validated QA        & \ding{51} & \ding{55} \\
        % News Fact Generation          & \ding{51} & \ding{51} \\
        % Retrieval-Augmented QA          & \ding{51} & \ding{51} \\
        
        Fact Sources & (1)~Human-written evidence ($S_{\text{he}}$); (2)~Reference documents ($S_{\text{rd}}$); (3)~Search engine results ($S_{\text{se}}$); (4)~LLM knowledge ($S_{\text{lk}}$).\\ \midrule
        Evaluation Scenarios &  
        (1)~$\langle S_{\text{se}}, S_{\text{lk}} \rangle$; 
(2)~$\langle S_{\text{he}}, S_{\text{se}}, S_{\text{lk}} \rangle$; 
(3)~$\langle S_{\text{rd}}, S_{\text{se}}, S_{\text{lk}} \rangle$; 
(4)~$\langle S_{\text{he}}, S_{\text{rd}}, S_{\text{se}}, S_{\text{lk}} \rangle$; 
(5)~$\langle S_{\text{rd}}, S_{\text{he}}, S_{\text{se}}, S_{\text{lk}} \rangle$.
        \\ \midrule
        Tasks & (1)~Open-domain QA; (2)~Web Retrieval-based QA; (3)~Expert-Validated QA; (4)~News Fact Generation; and (5)~Retrieval-Augmented QA.\\
    \bottomrule
    \end{tabular}
    \caption{The fact sources, evaluation scenarios, and tasks we study in the paper.}
    \label{tab:scenarios}
\end{table}

\section{Related Work}
\subsection{Text Generation and Hallucination}
The advancement of text generation has been propelled by pre-trained language models (PLMs) like BART~\cite{bart}, T5~\cite{T5}, and GPT-2~\cite{GPT2}, utilizing structures that range from encoder-decoder to decoder-only configurations. The emergence of LLMs such as GPT-3~\cite{GPT3}, characterized by their vast parameter counts and extensive training data, marked a significant evolution. These LLMs exhibit ``Emergent Abilities''~\cite{emergentAbility1} like In-Context Learning~\cite{ICL} and Chain-of-Thought Reasoning~\cite{emergentAbility2}. Despite these advancements, a challenge is the generation of text that deviates from human knowledge, known as \textit{hallucination}~\cite{Hallucination, faithfulnessSurvey}. Even the latest LLMs, such as GPT-4~\cite{GPT4}, still suffer from hallucinations, which greatly damages the factuality of the generated text.
% In the early years, the performance of text-generation tasks has been significantly enhanced by pre-trained language models (PLMs). These PLMs are based on the encoder-decoder structure, such as BART~\cite{bart} and T5~\cite{T5}, or the decoder-only structure, such as GPT-2~\cite{GPT2}, and apply self-supervised pre-training tasks on large-scale datasets. Recent studies have started to scale up the parameters of decoder-only PLMs as well as the amount of pre-training data. Such models are named large language models (LLMs), \eg, GPT-3~\cite{GPT3}. Researchers found that LLMs gain \textit{Emergent Abilities}~\cite{emergentAbility1, emergentAbility2}, such as In-Context Learning~\cite{ICL} and Chain-of-Thought~\cite{emergentAbility2}. However, the text generated by LLMs is often inconsistent with human knowledge, which is also known as \textit{hallucination}~\cite{Hallucination, faithfulnessSurvey}. Even the latest LLMs, such as GPT-4~\cite{GPT4}, still suffer from hallucinations, which greatly damages the factuality of the generated text.

In this paper, we propose a unified and flexible pipeline \ours to evaluate the factuality of the generated texts, which can detect hallucinations in various text generation tasks.

\subsection{Factuality Evaluation}

Factuality evaluation methods have evolved from traditional n-gram-based metrics to more sophisticated approaches leveraging PLMs and LLMs~\cite{faithfulnessSurvey}. 
Initially, metrics such as BLEU~\cite{bleu}, ROUGE~\cite{ROUGE}, and METEOR~\cite{meteor} assumed factual accuracy correlated with n-gram overlaps.
Later, metrics like BERTScore~\cite{Bertscore} utilizing contextual embeddings, and BARTScore~\cite{bartscore} employing generative scoring, captured deep semantic information between texts for evaluating factuality consistency. 
QAGS~\cite{QAGS} further innovates by combining entity extraction with PLM-based question generation and answering, while $Q^2$~\cite{Q2} leverages natural language inference (NLI) for entailment analysis. 
More recently, LLM-based metrics such as FactScore~\cite{factscore} and FacTool~\cite{factool} utilize LLM's reasoning ability, extracting and verifying facts against sources like Wikipedia dumps. 

Different from previous studies, our proposed pipeline \ours integrates human-written evidence, reference documents, search engine results, and LLM knowledge for factuality evaluation.

% \section{Task Design and Evaluation Framework}
\section{Methodology}
% 任务与目标
\subsection{Problem Statement}
Given a query $q_D$ sourced from a dataset $D$, an evaluated LLM $M$ generates a text passage $T_M(q_D)$. We define a collection of fact sources, denoted as $S$. The objective is to assign a factuality score $s \in [0, 1]$ to the model-generated text $T_M(q_D)$. A higher score denotes a greater consistency between the text $T_M(q_D)$ and the fact sources $S$, indicating higher factual accuracy of the LLM $M$.
% 事实源的定义和类型
\subsection{Fact Sources}
Based on the origin of fact sources, we categorize them into four types: human-written evidence ($S_{\text{he}})$, reference documents ($S_{\text{rd}}$), search engine results ($S_{\text{se}}$), and LLM knowledge ($S_{\text{lk}}$). Each type of fact source contains a series of text passages $\{P^1, P^2, \cdots\}$.
The first two types of fact sources ($S_{\text{he}}$ and $S_{\text{rd}}$) are provided by established datasets and require some cost to collect, such as responses and evidence written by users, and selected reference documents while they browse web pages.
The latter two ($S_{\text{se}}$ and $S_{\text{lk}}$) are fact sources relevant to specifically generated questions, independent of any particular dataset. These include text snippets retrieved from the web corpus and answers from the parameterized knowledge within LLMs.

For a given question, it might not be possible to obtain an answer from a certain fact source. Therefore, in an evaluation scenario, we predefine a sequence of fact sources $S = \langle S^1, S^2, \cdots \rangle$, and systematically verify each until a matched answer is extracted.

% UFO评价框架
\subsection{UFO Evaluation Framework}
% 整体介绍一下评价流程
Our evaluation pipeline includes three LLM-based modules: Fact Unit Extraction, Fact Source Verification, and Fact Consistency Discrimination. We employ OpenAI's ChatGPT API (\chatgpt) for these modules.
\begin{figure}[tb]
    \centering
    \includegraphics[width=1\linewidth]{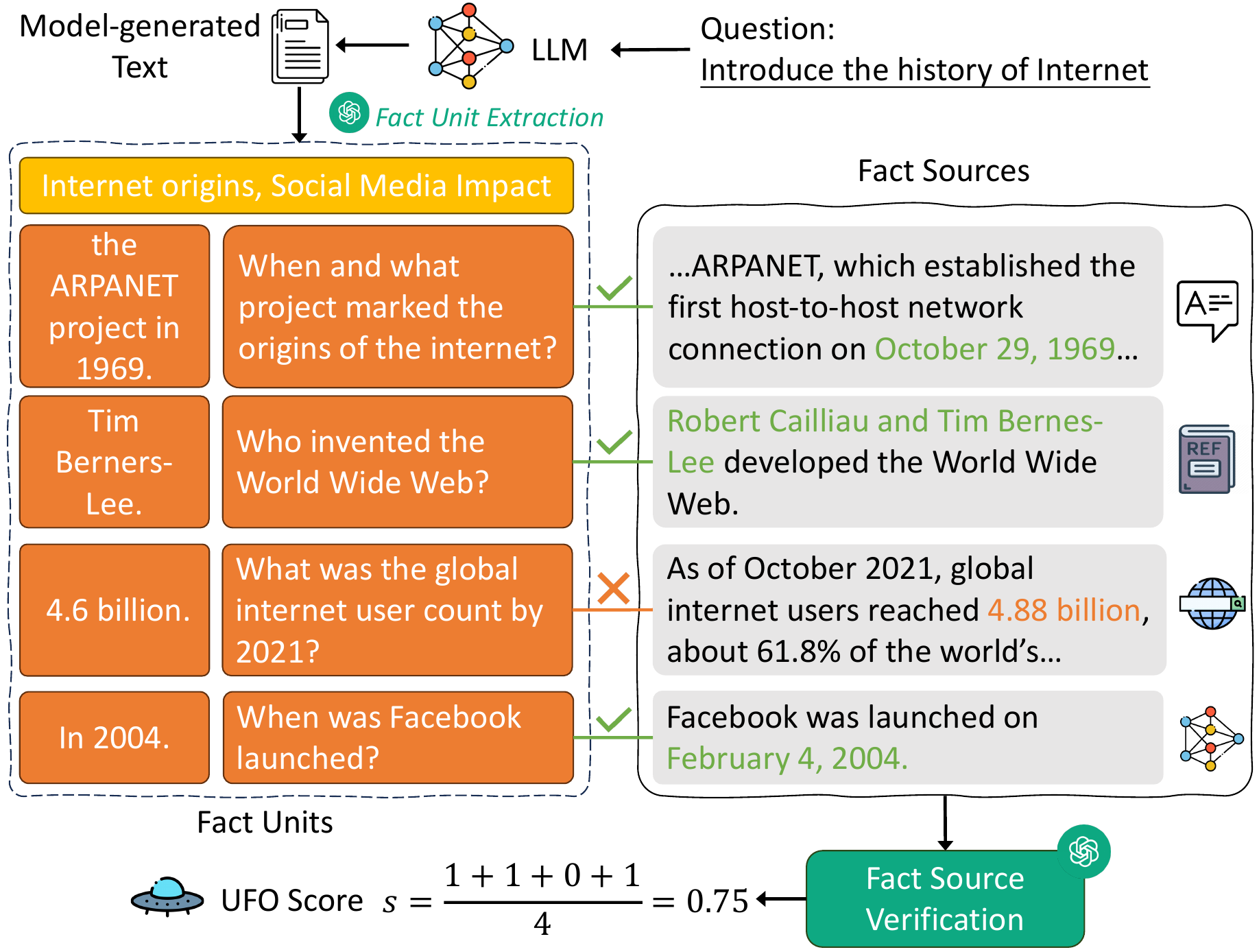}
    \caption{A case of evaluating Vicuna-generated text within the retrieval-augmented QA task where $S = \langle S_{\text{he}}, S_{\text{rd}}, S_{\text{se}}, S_{\text{lk}} \rangle$. Details of the generated text are omitted for clarity. The extracted answers are highlighted.}
    \label{fig:case}
\end{figure}
% 介绍具体的component
\subsubsection{Fact Unit Extraction}\label{FactUnitsExtraction}
LLMs can generate a text with several sentences for a given input, but not all the generated sentences are fact-related. Therefore, our first problem is to determine the smallest unit for factuality evaluation. 
We start by analyzing the process of factuality evaluation performed by humans. When faced with a text, humans will first focus on entities and their relevant descriptions that may cause factual errors. Then, they will ask a series of questions about the factuality of these descriptions. For example, when a text describes the date of birth $D$ of a famous person $X$, a common question is \textit{``when was $X$ born?''}. Finally, by comparing the golden answer $D'$ (from their knowledge or Internet) with $D$, the factuality of the description can be evaluated. 

Based on these analyses, we consider an entity-centric question $q_k$ and its corresponding answer $e_k$ can be used as a basic fact unit $f_k = \langle q_k, e_k \rangle$. 
% However, when verifying facts via a search engine, more open-ended questions lead to diverse search results. 
However, a generated question may have a weak relationship with its context. Therefore, we generate concise keywords of model-generated text.
For instance, a model-generated text is related to the demand for vinyl records in the Oxfam Charity Shop. the generated question, \textit{``What has led to a rise in demand for vinyl records?''} yields search engine results that discuss the recent global music trends. 
Thus, it is necessary to generate concise keywords $t$ for the model-generated text to refine the search engine's capability in retrieving content specifically relevant to the Oxfam Charity Shop.

Benefiting from the potent language comprehension capabilities of LLMs, we introduce an LLM-based Fact Unit Extraction (FUE) method to extract the fact units (Prompt~\ref{fig:factExtraction}) and generate the keywords (Prompt~\ref{fig:generate_title}) from the model-generated text. These prompts are provided in Appendix~\ref{prompt}.
\begin{align*}
    \{t, \langle q_1, e_1 \rangle, \cdots, \langle q_p, e_p \rangle \} = \text{FUE}(T_{M_i}(q_D)).
\end{align*}
Next, we will utilize the fact source sequence $S$ in different scenarios to evaluate the factual accuracy of these fact units.

\subsubsection{Fact Source Verification}
To verify the accuracy of a given fact unit $\langle q_i, e_i\rangle$ against the keywords $t$, our target is to identify the correct answer $a_i$ to the question $q_i$ using a specific text passage $P^k_j$ from a fact source $S^k$. 
However, not all text passages in the fact source are relevant to the question. To enhance the relevance of extracted answers, we employ the advanced language comprehension abilities of LLMs. 
We instruct the LLM-based Answer Extraction (AE) module to pinpoint the most relevant answers within the text, generating a ``[NOANS]'' text if no answer is found. 
This method involves directly prompting an LLM to retrieve answers from human-written evidence $S_{\text{he}}$ and reference documents $S_{\text{rd}}$ (Prompt~\ref{fig:AE}), reducing inaccuracies during fact verification~\cite{surveyHallucination}. 
When dealing with search engine results $S_{\text{se}}$ and LLM knowledge $S_{\text{lk}}$, we prompt the model to first check if an answer is available in the search results before resorting to its internal knowledge (Prompt~\ref{fig:llm+se}). 
Answers are sequentially sought in each text passage of the fact source until a suitable answer is found. If no text passage yields an answer, it indicates a mismatch with the fact source, leading to a transition to the next fact source $S^{k+1} \in S$ for verification.

Concretely, for a fact unit $\langle q_i, e_i \rangle$ and keywords $t$, we obtain the answer $a_i$ using passage $P^k_j$ from fact source $S^k$ as follows:
\begin{align}
     a_i &= \text{AE}(t, P^k_j, q_i), \\
     \text{AE}(t, P^k_m, q_i) &= \text{[NOANS]}, \\
     m &\in \{1, \cdots, j-1\}.
\end{align}

\subsubsection{Fact Consistency Discrimination}

Given the answer $e_i$ extracted from the model-generated text and the answer $a_i$ extracted from fact sources, our objective is to determine whether the two answers are factually consistent. To achieve this, we employ an LLM-based fact consistency discrimination (FCD) module (Prompt~\ref{fig:FCD}), assigning a score of 0 or 1 to each fact unit $\langle q_i, e_i \rangle$. Subsequently, we calculate the average score of all fact units as the factuality score of the model-generated text:
\begin{align}
    s_i &= \text{FCD}(e_i, a_i) \in \{0, 1\}, \\
    s &= \frac{1}{N}\sum_{i=1}^{N}s_i.
\end{align}

% \section{Application to Evaluation Scenarios}   
\subsection{Evaluation Criteria}

We measure the discriminative power (DP) of the evaluation metric, as described by~\citet{dp}. 
Given the collection of evaluated LLMs $M$ and all pairs $(M_i, M_j) \subset M$, we bootstrap sample the evaluation score on $M_i$ and $M_j$.
Then, given a threshold value $f$, we obtain minority rate (MR) and proportion of ties (PT) values.
The MR represents the failure rate of distinguishing the evaluation score differences between a pair of LLMs within the threshold.
The PT indicates the percentage of cases where the pair of LLMs cannot be distinguished within the threshold.
The smaller the values of MR and PT, the stronger the discriminative power of the evaluation metric.
Finally, we have the MR-PT curve as the discriminative power of the evaluation metric. 
The details of the pseudocode of DP measurement are given in Appendix~\ref{appendix:DP}.

\subsection{Evaluation Scenarios}
To assess the importance of each fact source across various tasks, we introduce five evaluation scenarios, each represented by an ordered list of fact sources $S$.
(1) $S = \langle S_{\text{se}}, S_{\text{lk}} \rangle$.
(2) $S = \langle S_{\text{he}}, S_{\text{se}}, S_{\text{lk}} \rangle$.
(3) $S = \langle S_{\text{rd}}, S_{\text{se}}, S_{\text{lk}} \rangle$.
(4) $S = \langle S_{\text{he}}, S_{\text{rd}}, S_{\text{se}}, S_{\text{lk}} \rangle$.
(5) $S = \langle S_{\text{rd}}, S_{\text{he}}, S_{\text{se}}, S_{\text{lk}} \rangle$.
To thoroughly verify facts in the text and mitigate the hallucination of LLM knowledge, we retain and fix the verification order of $S_{\text{se}}$ and $S_{\text{lk}}$.
By comparing the DP in scenarios (1), (2), and (3), we can infer the impact of fact sources. Comparing the DP in scenarios (4) and (5) reveals the effects of changing the verification order of fact sources.
% Based on the dataset-related fact sources $S_D \subset \{S_{\text{he}}, S_{\text{rd}}\}$), we categorize five typical text generation tasks into four types of evaluation scenarios (Table~\ref{tab:scenarios}):
% \textbf{(1) Open-domain QA}, addressing questions with potentially confusing facts, necessitating expert annotations for precision ($S_D = \varnothing$).
% \textbf{(2) Expert-validated QA}, tackling questions with potentially confusing facts, requiring clear expert annotations for accurate answers ($S_D = \{S_{\text{he}}\}$).
% \textbf{(3) Web Retrieval-based QA}, improving factual accuracy and entity relationships through web document retrieval ($S_D = \{S_{\text{rd}}\}$).
% \textbf{(4) News Fact Generation and Retrieval-Augmented QA}, utilizing both human-written evidence and reference documents for comprehensive answers ($S_D = \{S_{\text{he}}, S_{\text{rd}}\}$).
% In Figure~\ref{fig:case}, we demonstrate our proposed pipeline in the retrieval-augmented QA task where the fact sources in the scenario are denoted as $S = \langle S_{\text{he}}, S_{\text{rd}}, S_{\text{se}}, S_{\text{lk}} \rangle$.

Moreover, LLMs incorporating web search modules, such as Bing Chat, have been able to generate text while providing retrieved reference documents. In Section~\ref{sec:ModelDoc}, we will discuss the impact of using these referenced documents as the supplementary fact source $S_{\text{rd}}$ in evaluation scenarios.

\section{Experiments}

\subsection{Datasets}
Considering the available human-written evidence and reference documents, we categorize tasks presented in Table~\ref{tab:scenarios}. We carry out our evaluation pipeline on six datasets: NQ~\cite{NQopen}, HotpotQA~\cite{hotpotQA}, TruthfulQA~\cite{truthfulQA}, CNN/DM~\cite{cnndm}, Multi-News~\cite{multinews}, and MS MARCO~\cite{msmarco}.
We collect 200 samples from each dataset and prompt evaluated LLMs to generate verifiable facts in sufficient detail (Prompt~\ref{fig:genModelText}). 

To compare with reference-based metrics, we construct a golden answer $G$ containing more facts for each task.
\textbf{(1) Open-domain QA}: In the NQ dataset, we concatenated the provided short answers to form $G$.
\textbf{(2) Web Retrieval-based QA}: In the HotpotQA dataset, we combined the short answer and the reference documents as the golden answer $G = [a; S_{\text{rd}}]$.
\textbf{(3) Expert-validated QA}: In the TruthfulQA dataset, all provided human-written correct answers and best answers were considered as the fact source $S_{\text{he}}$, forming the golden answer $G$.
\textbf{(4) News Fact Generation}: For the CNN/DM and Multi-News datasets, we first prompted ChatGPT (Prompt~\ref{fig:generate_title_nfg}) to generate a title of the given summary (considered as $S_{\text{he}}$). Then, we used the generated title to prompt the evaluated LLMs (Prompt~\ref{fig:genModelText}) to generate an introduction centered around the facts.
\textbf{(5) Retrieval-Augmented QA}: In the MS MARCO dataset, the answer $a$ was regarded as $S_{\text{he}}$, and all user-clicked documents were considered as $S_{\text{rd}}$. The answer and the selected documents were concatenated to form $G$.

\subsection{Baselines}
% \begin{table*}[tb]
% \centering
% \small
% \caption{The average proportion of facts verified by \textbf{human-written evidence} and \textbf{reference documents} in the fact source verification under different fact source settings.}
% \begin{center}
% \begin{tabular}{lrrrrrr}
% \toprule
% \multirow{3}{*}{Fact Source Setting} & \multicolumn{6}{c}{Dataset}\\
% \cmidrule(lr){2-7}
%  & NQ & HotpotQA & TruthfulQA & CNN/DM & Multi-News & MS MARCO \\     
% \midrule
% $\{S_{\text{sr}}\}$                    & -~/~- & -~/~- & -~/~- & -~/~- & -~/~- & -~/~- \\
% $\{S_{\text{he}}, S_{\text{sr}}\}$            & -~/~- & -~/~- & 64\%~/~- & 73\%~/~- & 76\%~/~- & 48\%~/~- \\
% $\{S_{\text{rd}}, S_{\text{sr}}\}$            & -~/~- & -~/~72\% & -~/~- & -~/~79\% & -~/~72\% & -~/~50\% \\
% $\{S_{\text{he}}, S_{\text{rd}}, S_{\text{sr}}\}$    & -~/~- & -~/~72\% & 64\%~/~- & 73\%~/~14\% & 76\%~/~6\% & 48\%~/~4\%\\
% $\{S_{\text{rd}}, S_{\text{he}}, S_{\text{sr}}\}$    & -~/~- & -~/~72\% & 64\%~/~- & 8\%~/~79\% & 10\%~/~72\% & 2\%~/~48\% \\
% \bottomrule
% \end{tabular}
% \end{center}
% \end{table*}
\begin{table}[tbp]
\centering
\scriptsize
\begin{center}
\begin{tabular}{lrrrrrr}
\toprule
$S_{\text{he}}$ & \ding{55} & \ding{55} & \ding{51} & \ding{51} & \ding{51} & \ding{51} \\       
$S_{\text{rd}}$ & \ding{55} & \ding{51} & \ding{55} & \ding{51} & \ding{51} & \ding{51} \\
Dataset  &  NQ  & HQA  &  TQA  &  C/D  & 
 M-N  &  MS \\
\midrule
\multicolumn{7}{l}{Avg. \# of Tokens} \\
\midrule
Bing Chat & 136.96 & 87.99 & 196.02 & 223.64 & 248.66 & 287.93 \\
ChatGPT  & 398.36  & 336.26   & 393.41     & 534.74 & 531.17  & 561.91  \\ 
llama-7b & 455.14  & 427.46   & 453.58 & 433.66 & 436.47 & 459.84 \\
llama-13b & 431.99  & 406.42 & 432.24 & 422.61 & 424.83 & 455.64 \\
vicuna-7b & 353.72 & 332.16 & 387.88 & 398.84 & 401.92 & 413.28 \\ 
vicuna-13b & 341.00 & 327.38 & 346.32 & 366.60 & 369.31 & 398.76 \\ 
\midrule
\multicolumn{7}{l}{Avg. \# of Sentences} \\
\midrule
Bing Chat & 5.15 & 3.42 & 7.50 & 8.00 & 8.20 & 10.54  \\
ChatGPT   & 12.46 & 10.05 & 12.62 & 15.47 & 15.57 & 18.66   \\ 
llama-7b  & 17.34 & 15.80  & 18.41 & 15.73 & 16.21 & 18.85  \\
llama-13b & 15.88 & 14.54 & 16.70 & 14.64 & 14.86 & 17.19  \\
vicuna-7b & 12.25 & 11.39  & 14.50 & 13.03 & 13.17 & 15.98  \\ 
vicuna-13b& 11.72 & 10.85 & 12.31 & 12.16 & 12.10 & 14.77 \\ 
\midrule
\multicolumn{7}{l}{Avg. \# of Facts Extracted Using \textit{ChatGPT}}\\
\midrule
Bing Chat & 4.12 & 3.08 & 4.78 & 5.32 & 5.64 & 5.85 \\
ChatGPT   & 5.60 & 5.17 & 5.41 & 5.94 & 5.92 & 6.04 \\
llama-7b  & 4.86 & 4.97 & 5.06 & 5.15 & 5.18 & 5.30 \\
llama-13b & 5.14 & 5.11 & 5.00 & 4.96 & 5.28 & 5.11 \\
vicuna-7b & 5.76 & 5.66 & 5.34 & 6.04 & 6.18 & 5.46 \\
vicuna-13b& 5.51 & 5.56 & 5.26 & 5.90 & 5.66 & 5.62 \\
\bottomrule
\label{stat}
\end{tabular}
\caption{Statistics of model-generated text on six datasets. ``HQA'', ``TQA'', ``C/D'', ``M-N'', and ``MS'' are abbreviations of ``HotpotQA'', ``TruthfulQA'', ``CNN/DM'', ``Multi-News'' and ``MS MARCO''.}
\label{tab:stat}
\end{center}
\end{table}

\noindent\textbf{Evaluated Models}\quad
We evaluate six existing LLMs with varying parameter scales in our experiments:
(1) Bing Chat is a GPT-4-based model specifically tailored for web searches. For this model, we choose the ``Precise'' generation mode to test the factuality when the model is expected to generate the most accurate and detailed fact units.\footnote{\url{https://blogs.bing.com/search/march_2023/Confirmed-the-new-Bing-runs-on-OpenAI\%E2\%80\%99s-GPT-4}}
In each provided URL, we extract all the <p> tags of the corresponding web page. Subsequently, we divide the text into multiple passages, each containing no more than 1024 tokens.
(2) ChatGPT: we utilized OpenAI's ChatGPT API (\chatgpt) for text generation.\footnote{\url{https://platform.openai.com/docs/api-reference/chat}}
(3) LLaMA~\cite{llama}: We select two LLaMA-series fine-tuned models for text generation (LLaMA-2-7b-chat and LLaMA-2-13b-chat).
(4) Vicuna~\cite{vicuna}: Vicuna is a chat assistant developed by fine-tuning LLaMA-2 foundation model with user-shared conversations collected from ShareGPT.\footnote{\url{https://sharegpt.com/}} We select Vicuna-7b-v1.5 and Vicuna-13b-v1.5 to generate text. 
The statistical data of the text generated by these LLMs is given in Table~\ref{tab:stat}.

\noindent\textbf{Baseline Evaluation Metrics}\quad
We compare our proposed pipeline with both reference-based and reference-free metrics.

\textbf{(1) Reference-based metrics.} Such metrics require a golden answer $G$ and calculate the consistency with the model-generated text.
BLEU~\cite{bleu} and ROUGE~\cite{ROUGE} are used to measure the token-level term overlap. BERTScore~\cite{Bertscore} and BARTScore~\cite{bartscore} are model-based metrics to evaluate passage-level similarity. QAGS~\cite{QAGS} and $Q^2$~\cite{Q2} are the most relevant PLM-based and NLI-based metrics to evaluate factuality.

\textbf{(2) Reference-free metrics.} FactScore~\cite{factscore} first breaks down the model-generated text into several claims. Subsequently, these claims are verified through Wikipedia dumps. In this study, we form all golden answers as the corpus for FactScore verification. FacTool~\cite{factool} performs the verification of each claim by employing a search engine and derives factuality scores at the claim level.

\begin{figure*}[htbp]
    \centering
    \includegraphics[width=1\linewidth]{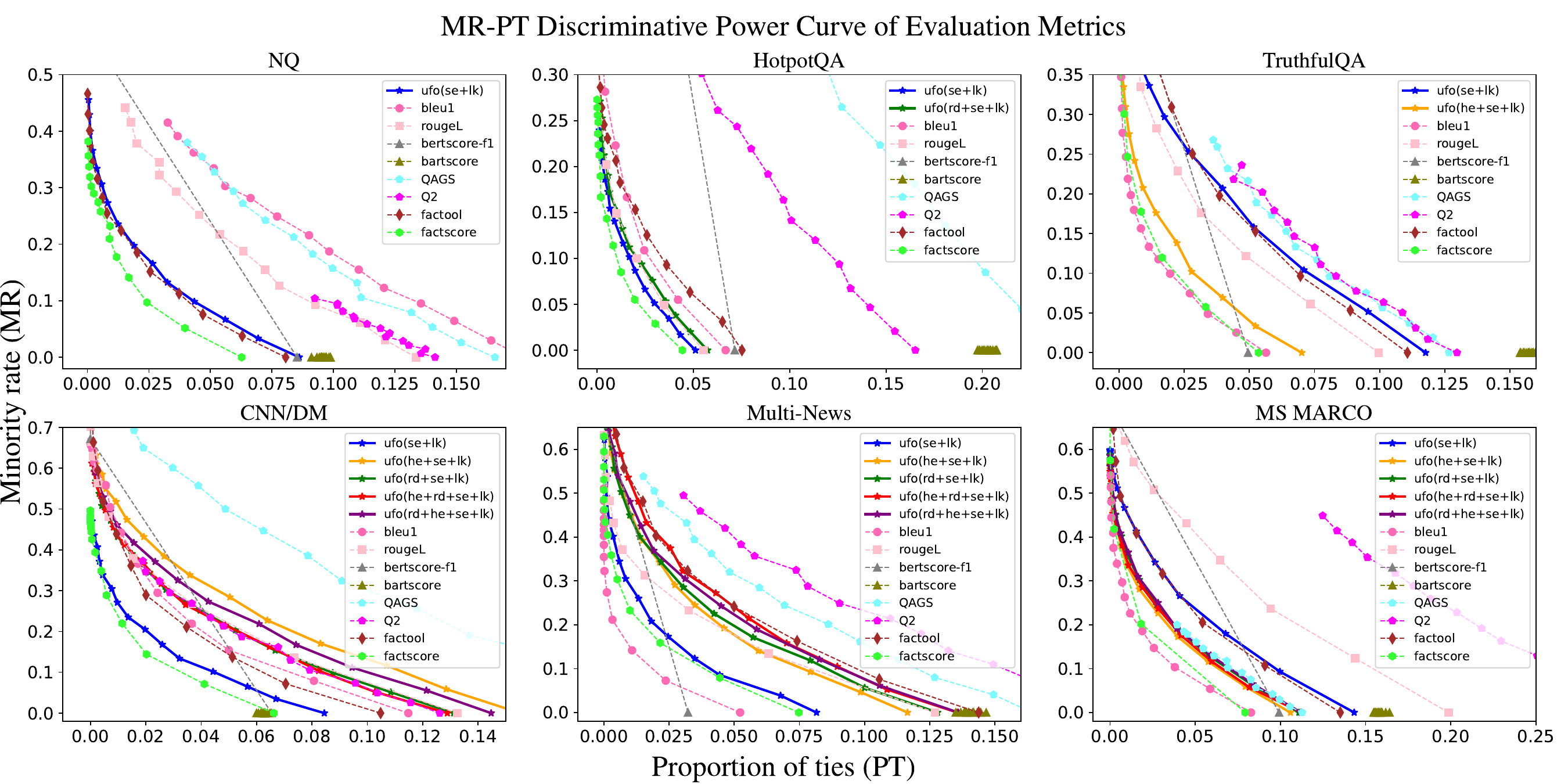}
    \caption{MR-PT discriminative power curve of evaluation metrics on datasets. 
    % The discriminative power of the evaluation metric increases as its curve approaches the origin of the coordinate axis. 
    The closer the curve is to the bottom-left corner, the better the evaluation metric is.
    Our proposed model curve is represented by a \textbf{solid line}, while the baseline model curve is depicted using a \textbf{dashed line}.}
    \label{fig:dp}
\end{figure*}
\begin{figure*}[htbp]
    \centering
    \includegraphics[width=1\linewidth]{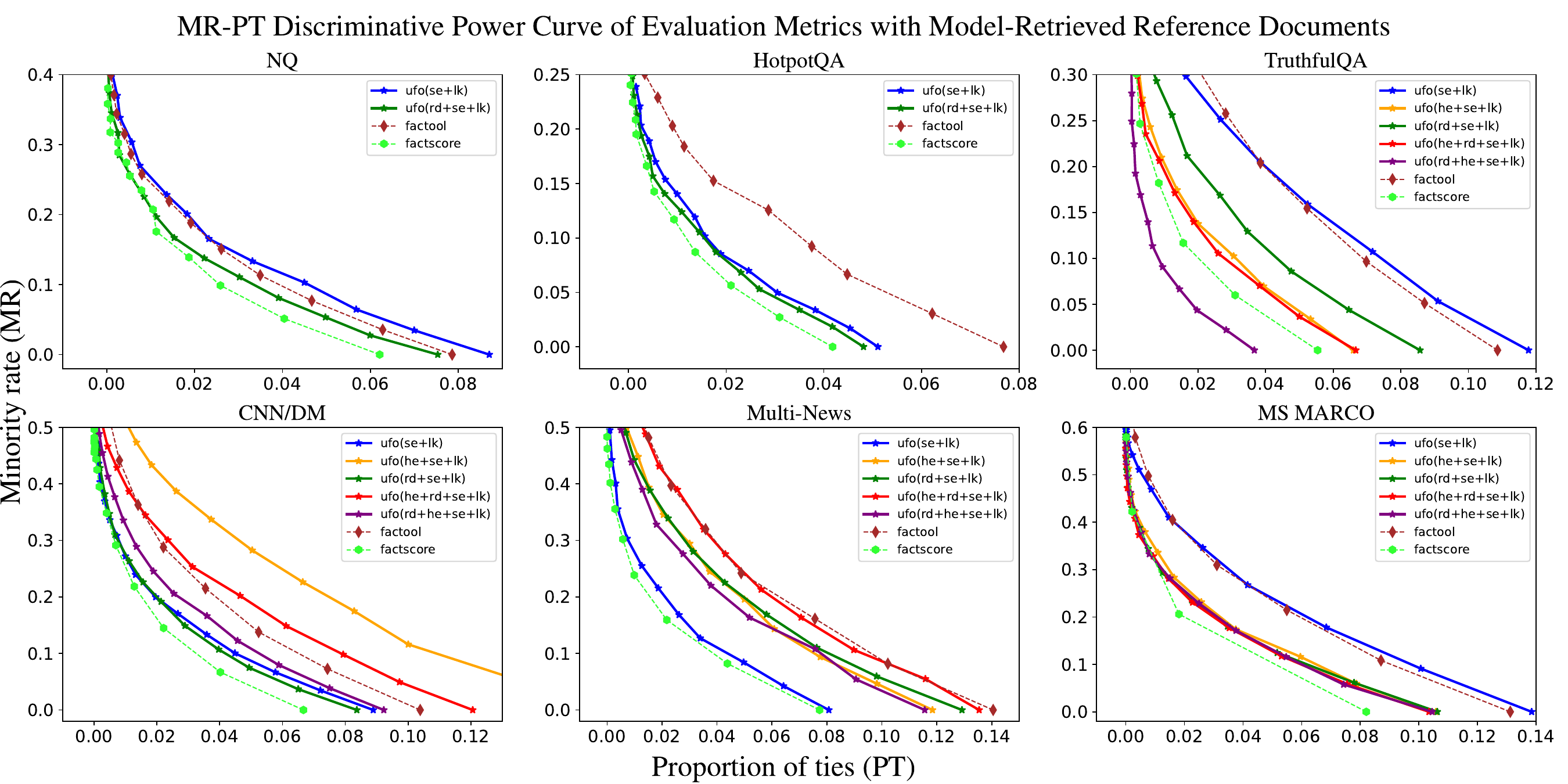}
    \caption{MR-PT discriminative power curve of evaluation metrics on datasets. The closer the curve is to the bottom-left corner, the better the evaluation metric is. We incorporate reference documents retrieved by Bing Chat as part of the fact source $S_{\text{rd}}$. Non-LLM-based methods are omitted for clarity.}
    \label{fig:dp-useModelDoc}
\end{figure*}
\begin{table*}[htbp]
\centering
\small
\begin{center}
\begin{tabular}{lcccccccc}
\toprule
\multirow{3}{*}{Scenarios} &  \multicolumn{2}{c}{Pearson $\uparrow$} & \multicolumn{2}{c}{Spearman $\uparrow$} & \multicolumn{2}{c}{\#Avg. Tokens $\downarrow$} & \multicolumn{2}{c}{\#Avg. Time $\downarrow$}\\
\cmidrule(lr){2-3}\cmidrule(lr){4-5}\cmidrule(lr){6-7}\cmidrule(lr){8-9}
 & FacTool & FactScore & FacTool & FactScore & FacTool & FactScore & FacTool & FactScore \\     
\midrule
$\langle S_{\text{se}}, S_{\text{lk}} \rangle$         & \textbf{0.296} & 0.272  & \textbf{0.308} & 0.283 & +1.20\% & -6.71\% & \textbf{-3.44\%} & \textbf{-22.31\%} \\
$\langle S_{\text{he}}, S_{\text{se}}, S_{\text{lk}} \rangle$ & 0.275 & 0.269 & 0.276 & 0.270 & \textbf{-3.79\%} & \textbf{-12.21\%} & -2.03\% & -19.26\% \\
$\langle S_{\text{rd}}, S_{\text{se}}, S_{\text{lk}} \rangle$         & 0.279 & 0.278 & 0.283 & 0.290 & +12.33\% & -9.97\% & -2.31\% & -16.02\% \\
$\langle S_{\text{he}}, S_{\text{rd}}, S_{\text{se}}, S_{\text{lk}} \rangle$   & 0.263 & 0.269 & 0.271 & 0.281 & +6.25\% & -11.03\% & +2.43\% & -18.75\% \\
$\langle S_{\text{rd}}, S_{\text{he}}, S_{\text{se}}, S_{\text{lk}} \rangle$   & 0.267 & \textbf{0.283} & 0.271 & \textbf{0.295} & +11.20\% & -10.35\% & +7.13\% & -19.63\% \\
\bottomrule
\end{tabular}
\caption{
Pearson's and Spearman's correlation coefficients for 
\ours and the LLM-based baseline evaluation metrics under five evaluation scenarios. All $p$-values are less than $0.01$. Additionally, we compared the usage of ChatGPT API tokens and the average time required for evaluating samples. The best results are marked \textbf{bold}.}
\label{tab:correlation}
\end{center}
\end{table*}
\section{Results and Analysis}
\subsection{Discriminative Power Results}\label{sec:dp}
Our goal is to evaluate the discriminative power~\cite{dp} of the proposed evaluation pipeline \ours in each scenario. 
For simplicity, in the scenario such as $S = \langle S_{\text{se}}, S_{\text{lk}} \rangle$, we name our framework ``ufo(se+lk)''.
The experimental results are shown in Figure~\ref{fig:dp}. 
% The horizontal axis represents the proportion of samples that could not be distinguished by the evaluated model, while the vertical axis indicates the overall proportion of samples correctly differentiated. 
Each point on the curve represents the values of MR and PT calculated at a given threshold $f$.
We have the following findings:

(1) Among baseline metrics, BARTScore demonstrates minimal variance with a notably low MR value, while QA-based metrics like QAGS and $Q^2$ show suboptimal discriminative power across all tasks. It indicates that PLM-based methods are particularly reliant on the quality of the golden answer, especially its entities and relationships.
FactScore~\cite{factscore} verifies each extracted claim against a predefined fact source and enhances the LLM-based method's capability through In-Context Learning with demonstrations. Thus it shows relatively high discriminative power with additional time and token usage. We will discuss the API and time usage in Section~\ref{sec:usage}.

% (3) In the NQ dataset, FacTool and our proposed method \ours demonstrate comparable discriminative power, suggesting that the LLM-based methods for extracting QA pairs and generating claims are comparable.

(2) HotpotQA and TruthfulQA dataset provide $S_{\text{rd}}$ and $S_{\text{he}}$ respectively. However, the proposed pipeline \ours in the scenario $\langle S_{\text{rd}}, S_{\text{se}}, S_{\text{lk}} \rangle$ in HotpotQA resulted in weaker DP. It suggests that the quality of $S_{\text{rd}}$ is inferior to search engine results $S_{\text{se}}$ and LLM knowledge $S_{\text{lk}}$. Search engines can retrieve more relevant details for entities involved in multi-hop reasoning. In TruthfulQA, there is a significant presence of questions with confusion, hence the fact source $S_{\text{he}}$ has higher quality. \ours in the scenario $\langle S_{\text{he}}, S_{\text{se}}, S_{\text{lk}} \rangle$ show a substantial increase in DP compared to the scenario without $S_{\text{he}}$. This also indicates the necessity of incorporating human-written evidence in expert-validated QA tasks.

(3) CNN/DM, Multi-News, and MS MARCO provide both $S_{\text{he}}$ and $S_{\text{rd}}$. However, in the task of news fact generation, \ours in the scenario $S = \langle S_{\text{se}}, S_{\text{lk}}\rangle$ achieves the highest DP, indicating that both $S_{\text{he}}$ and $S_{\text{rd}}$ exhibit a negative impact on DP. In fact, within provided documents and summaries, the factual details are considerably limited and not specific enough. Thus, employing search engines enables precise retrieval of factual details based on specific extracted questions. 
In the retrieval-augmented QA task, $S_{\text{he}}$ and $S_{\text{rd}}$ significantly enhance DP. Moreover, changing the order of verification between $S_{\text{he}}$ and $S_{\text{rd}}$ does not significantly affect the DP value. This reflects a high degree of consistency between user-clicked reference documents and human-written evidence. 
It suggests that for this task, reference documents and human-written evidence can be substituted for each other, and we can rely solely on user-clicked reference documents without the need for collecting human-written evidence.
\begin{table*}[tp]
\centering
\small
\begin{center}
\begin{tabular}{lrrrrrrrrrrrr}
\toprule
% $S_{\text{he}}$ & \multicolumn{2}{c}{\ding{55}} & \multicolumn{2}{c}{\ding{55}} & \multicolumn{2}{c}{\ding{51}} & \multicolumn{2}{c}{\ding{51}} & \multicolumn{2}{c}{\ding{51}} & \multicolumn{2}{c}{\ding{51}} \\
% $S_{\text{rd}}$ & \multicolumn{2}{c}{\ding{55}} & \multicolumn{2}{c}{\ding{51}} & \multicolumn{2}{c}{\ding{55}} & \multicolumn{2}{c}{\ding{51}} & \multicolumn{2}{c}{\ding{51}} & \multicolumn{2}{c}{\ding{51}} \\
Dataset  &  \multicolumn{2}{c}{NQ}  & \multicolumn{2}{c}{HotpotQA}  &  \multicolumn{2}{c}{TruthfulQA}  &  \multicolumn{2}{c}{CNN/DM}  & 
 \multicolumn{2}{c}{Multi-News}  &  \multicolumn{2}{c}{MS MARCO} \\
\cmidrule(lr){2-3}\cmidrule(lr){4-5}\cmidrule(lr){6-7}\cmidrule(lr){8-9}\cmidrule(lr){10-11}\cmidrule(lr){12-13}
Metrics & UFO & FT & UFO & FT & UFO & FT & UFO & FT & UFO & FT & UFO & FT\\
\midrule
% Bing Chat & & 0.535 & & 0.420 & & 0.616 & & 0.670 & & 0.688 & & 0.740 \\
Bing Chat & \underline{0.752} & 0.615 & \underline{0.630} & \underline{0.709} & 0.649 & 0.594 & 0.628 & \underline{0.742} & \underline{0.685} & \underline{0.745} & 0.725 & \underline{0.787}\\
ChatGPT   & \textbf{0.762} & \textbf{0.776} & \textbf{0.635} & \textbf{0.725} & 0.662 & \textbf{0.700} & \textbf{0.669} & \textbf{0.806} & \textbf{0.708} & \textbf{0.806} & \textbf{0.765} & \textbf{0.845}\\
llama-7b  & 0.610 & 0.480 & 0.465 & 0.453 & 0.630 & 0.560 & 0.607 & 0.673 & 0.589 & 0.689 & 0.711 & 0.757\\
llama-13b & 0.674 & 0.596 & 0.537 & 0.537 & 0.597 & 0.564 & 0.573 & 0.688 & 0.661 & 0.731 & 0.735 & 0.731\\
vicuna-7b & 0.670 & 0.631 & 0.515 & 0.562 & \underline{0.692} & 0.610 & 0.603 & 0.714 & 0.593 & 0.688 & 0.662 & 0.755\\
vicuna-13b& 0.676 & \underline{0.658} & 0.514 & 0.570 & \textbf{0.717} & \underline{0.664} & \underline{0.652} & 0.740 & 0.646 & 0.731 & \underline{0.739} & 0.778\\
\bottomrule
\end{tabular}
\caption{Factuality scores of our proposed evaluation framework \ours in the scenario of $S = \langle S_{\text{he}}, S_{\text{rd}}, S_{\text{se}}, S_{\text{lk}} \rangle$ and FacTool (abbreviated to ``FT'') on six datasets. In the evaluation of the group of evaluated LLMs, the highest factuality score is \textbf{bold}, and the second highest score is \underline{underlined}.}
\label{tab:fact_score}
\end{center}
\end{table*}
\subsection{Effect of Model-Retrieved Documents}\label{sec:ModelDoc}
Some existing LLMs provide retrieved reference documents during text generation. We incorporate these as part of $S_{\text{rd}}$ to evaluate the certain LLM (\ie, Bing Chat in our experiments). The discriminative power of LLM-based evaluation metrics is shown in Figure~\ref{fig:dp-useModelDoc}. We have the following findings:

(1) In the NQ, HotpotQA, and TruthfulQA datasets, incorporating retrieved reference documents in the evaluation scenarios enhances the DP. Specifically on the TruthfulQA dataset, we further observe that \ours in the scenario $\langle S_{\text{rd}}, S_{\text{he}}, S_{\text{se}}, S_{\text{lk}} \rangle$ significantly boosts discriminative power and surpasses FactScore. This implies that the model accurately retrieves relevant reference documents based on easily confused facts during text generation.

(2) In the CNN/DM dataset, \ours in the scenario $\langle S_{\text{rd}}, S_{\text{se}}, S_{\text{lk}} \rangle$ shows a slight improvement when incorporating model-retrieved documents compared to the use of search engines and LLM knowledge $\langle S_{\text{se}}, S_{\text{lk}}\rangle$. It suggests that retrieved reference documents serve as a beneficial complement to search engine results in this task.

(3) In the MS MARCO dataset, the enhancement of DP brought by incorporating retrieved reference documents is minimal, indicating a factual consistency between human-written evidence, clicked reference documents, and retrieved reference documents.

\subsection{Factuality Scores of LLMs}
In addition to evaluating discriminative power, we also calculated the factuality scores of evaluated LLMs on various datasets.
Under the evaluation scenario $S = \langle S_{\text{he}}, S_{\text{rd}}, S_{\text{se}}, S_{\text{lk}} \rangle$, the comparative experimental results between our proposed framework \ours and FacTool are presented in Table~\ref{tab:fact_score}. 
Both evaluation methods show that in most datasets, the factuality score of Bing Chat in ``precise'' mode is slightly lower than that of ChatGPT, but close to the score of Vicuna-13b. 
This implies that hallucinations occur during the retrieval-augmented generation process, thereby reducing the factual accuracy of the generated text.
We also observe that increasing the parameter scale of open-source LLMs (LLaMA and Vicuna) can enhance factual accuracy in most datasets.

% \subsection{Verification Proportion of Fact Sources}\label{sec:proportion}

\subsection{Cost of LLM-based Metrics}\label{sec:usage}
Our proposed pipeline sequentially extracts answers from listed text passages in the fact source. To assess the efficiency of our proposed evaluation pipeline, we compared the average evaluation time, API token costs, and the correlation coefficient with existing LLM-based evaluation metrics. Experimental results (shown in Table~\ref{tab:correlation}) demonstrate that our proposed pipeline achieves the highest correlation coefficient with FacTool and FactScore in the scenarios where the fact sources are set to $\langle S_{\text{se}}, S_{\text{lk}} \rangle$ and $\langle S_{\text{rd}}, S_{\text{he}}, S_{\text{se}}, S_{\text{lk}} \rangle$ respectively. 

Besides, in comparison to FactScore, our proposed evaluation pipeline reduces token consumption by about 10\% and time cost by about 20\%,
% Besides, the evaluation time and API costs of our proposed pipeline are less than those of FactScore, 
while our proposed pipeline maintains a relatively comparable discriminative power. Compared to FacTool, in most five evaluation scenarios, we achieved greater discriminative power with an affordable additional cost of about 10\% in all tasks. It implies that the incorporation of fact sources can enhance the discriminative power of evaluation metrics.

\section{Conclusion}
In this paper, we propose \texttt{UFO}, a factuality evaluation pipeline incorporating flexible plug-and-play fact sources: human-written evidence, reference documents, search engine results, and LLM knowledge with unified verification methods.
Experimental results on five evaluation scenarios show that open-domain QA, web retrieval-based QA, and expert-validated QA tasks require high-quality human-written evidence and model-retrieved reference documents, and retrieval-augmented QA needs either human-written evidence or user-clicked reference documents, while news fact generation tasks rely on search engine results and LLM knowledge.

%TODO Comment this out
% \clearpage

\section*{Limitations}
Though our proposed pipeline analyzes different fact sources, there are still several limitations:
% 依赖大模型的质量
(1) We utilize ChatGPT (\chatgpt) in the modules of our proposed evaluation framework, therefore the updating of the LLM will affect both the cost and effectiveness of our evaluation.
% 我们的方法还不能和training结合在一起，怎么以强化学习的方式辅助训练
(2) Currently, our approach has not yet been integrated with the training process of LLMs. In future work, we will consider incorporating the factuality score evaluated by our framework into the training process of the LLMs through reinforcement learning methods.

\section*{Ethics Statement}
The datasets used in this paper are available publicly online. Particularly, any data involving sensitive information has been anonymized, ensuring that it cannot be traced back to individuals.
% \bibliography{anthology,custom}
\bibliography{custom}
\clearpage
\appendix
\section{Prompt}\label{prompt}
\begin{prompt}[title={Prompt 1: Fact unit extraction using ChatGPT}, label=fig:factExtraction]
Your task is to segment a given document into several atomic claims. For each claim, you need to generate several questions related to it and extract an answer for each question from that claim. Your output is a JSON list. Each element includes the question, the answer, and the sentence from the document containing the atomic claims.

You MUST only respond in the JSON List format as described below. DO NOT RESPOND WITH ANYTHING ELSE. ADDING ANY OTHER EXTRA NOTES THAT VIOLATE THE RESPONSE FORMAT IS BANNED. START YOUR RESPONSE WITH ’[’.
[Response Format]
[{"question": "Informative question", "answer": "A concise phrase under 10 words", "sentence": "Sentence containing the answer."}, ...]

document: \{document\}
\end{prompt}
\begin{prompt}[title={Prompt 2: Generate keywords for model-generated text}, label=fig:generate_title]
Generate keywords for the following document. Do not provide any explanations.

document: \{document\}

keywords:
\end{prompt}
\begin{prompt}[title={Prompt 3: LLM-based answer extraction with $S_{\text{he}}$ and $S_{\text{rd}}$}, label=fig:AE]
You are an answer-extraction expert. Your task is to extract a short answer from the evidence to the question. Directly answer without any explanations. If the evidence is irrelevant to the question, respond ONLY with "NOANS".

keywords: \{keywords\}

evidence: \{evidence\}

question: \{question\}

your answer:
\end{prompt}
\begin{prompt}[title={Prompt 4: LLM-based answer extraction with $S_{\text{se}}$ and $S_{\text{lk}}$}, label=fig:llm+se]
You are a question-answering expert. You are given a question, keywords, and some snippets. Your task is to output a short answer to the question based on the snippets or the knowledge you possess, while your answer is factually consistent with the given keywords. If your answer is based on the snippets, you should provide the indices of the snippets. If there is no relevant snippet, you should answer with the knowledge you possess, and the output index is [-1]. If you are uncertain about the correctness and timeliness of your answer, your answer should be formed as [NOANS] instead. An example output format: [<your answer>]; [<index1>, <index2>, …]. Your output MUST begin with ‘[‘. DO NOT GIVE ANY EXPLANATIONS.

question: \{question\}

keywords: \{keywords\}

snippets: \{snippets\}
\end{prompt}

\begin{prompt}[title={Prompt 5: LLM-based fact consistency discrimination}, label=fig:FCD]
Your task is to judge whether the following two answers are factually consistent. Directly answer yes or no.

Answer 1: \{$e_i$\}

Answer 2: \{$a_i$\}
\end{prompt}

\begin{prompt}[title={Prompt 6: Generate long-form text}, label=fig:genModelText]
You have been presented with the following title. Your task is to provide a comprehensive introduction to the query topic with sufficient verifiable facts based on the knowledge you possess. Your output must be in English.

Title: \{title\}

Introduction:
\end{prompt}
\begin{prompt}[title={Prompt 7: Generate title for model-generated text}, label=fig:generate_title_nfg]
Generate a summarized title for the following document. Do not provide any explanations.

document: \{document\}

title:
\end{prompt}
\section{Pseudocode for DP Measurement}\label{appendix:DP}
\begin{algorithm}[h]
\caption{Discriminative Power Measurement with MR-PT Curve} % 算法的标题
\label{alg:dp_algo}
\begin{algorithmic}[1] % 数字1表示每行都显示数字
\For{\textbf{each} $threshold \in \{0, 0.01, \cdots, 0.20\}$}
    \State $f \gets threshold$
    \State $B \gets 1000$
    \For{\textbf{each} $(M_i, M_j) \in M$} % For循环
        \For{$b=1$ to $B$} % For循环
            \State $Q_i = mean(Bootstrap(M_i))$
            \State $Q_j = mean(Bootstrap(M_j))$
            \State $m=f*\text{max}(Q_i, Q_j)$
            \If{$|Q_i - Q_j| < m$}
                \State $EQ(i, j) \gets EQ(i, j) + 1$
            \ElsIf{$Q_i > Q_j$}
                \State $GT(i, j) \gets GT(i, j) + 1$
            \Else
                \State $GT(j, i) \gets GT(j, i) + 1$
            \EndIf
        \EndFor
    \EndFor
    \State $MR_f \gets \frac{\sum_{M_i, M_j}\text{min}(GT(i, j), GT(j, i))}{B\sum_{M_i, M_j}}$
    \State $PT_f \gets \frac{\sum_{M_i, M_j}EQ(i, j)}{B\sum_{M_i, M_j}}$
\EndFor
\end{algorithmic}
\end{algorithm}
\end{document}